\documentclass{article}

\usepackage{iclr2026_conference,times}

\usepackage{amsmath,amsfonts,bm}

\def\eqref#1{equation~\ref{#1}}

\def\1{\bm{1}}

\def\F2{{$\mathbb{F}_2$}}

\DeclareMathAlphabet{\mathsfit}{\encodingdefault}{\sfdefault}{m}{sl}
\SetMathAlphabet{\mathsfit}{bold}{\encodingdefault}{\sfdefault}{bx}{n}

\usepackage{amsmath,amssymb,amsthm}
\usepackage{graphicx}
\usepackage{import}
\usepackage{hyperref}
\usepackage{booktabs}

\title{Controlling Logical Collapse in LLMs via Algebraic Ontology Projection over \F2}

\author{Hisashi Miyashita \\
Mgnite Inc.\\
\texttt{himi@mgnite.com}
}

\iclrfinalcopy %
\begin{document}

\maketitle

\begin{abstract}
Do large language models internally encode ontological
relations in a formally verifiable algebraic structure?
We introduce \textbf{Algebraic Ontology Projection~(AOP)},
which projects LLM hidden states into the Galois
Field~$\mathbb{F}_2$ under Liskov Substitution Principle
constraints, using only 42~relational pairs as algebraic
keys.
AOP achieves up to \textbf{93.33\%} zero-shot inclusion
accuracy on unseen concept pairs~(Gemma-2 Instruct with
optimized prompt), with consistent \textbf{86.67\%}
accuracy observed across multiple model families---
with no model tuning, but through prompt alone.

This algebraic structure is strongly layer-dependent.
We introduce \textbf{Semantic Crystallisation~(SC)}, a
metric that quantifies $\mathbb{F}_2$ constraint
satisfaction relative to a random baseline and predicts
zero-shot accuracy without held-out data.
System prompts act as \textbf{algebraic boundary
conditions}: only their combination with instruction
tuning prevents \textbf{Late-layer Collapse}---a
systematic degradation of logical consistency in the
final layers, observed in 7~of~10~conditions.

These findings reframe forward computation as an iterative
process of algebraic organisation, and open a path toward
LLMs whose logical structure is not merely approximated,
but formally accessible.
\end{abstract}

\section{Introduction}
\label{sec:intro}

When a large language model correctly identifies that an Eagle
\textit{is a} Bird, that a Salmon \textit{is a} Fish, and that
a Diamond is \textit{not} an Insect, is this behaviour the
product of geometric proximity in embedding space---or does the
model internally encode something more structured?

We provide evidence for the latter.
Using a projection of LLM hidden states onto a binary vector
space under hierarchical constraints, we show that relational
structure---\textit{is-a}, \textit{has-a}, and negation---is
not merely approximated geometrically, but encoded in a form
that is extractable by linear means, and that generalises
to concepts never seen during projection.

\begin{figure}[h]
 \centering
 \def\svgwidth{0.9\linewidth}
 \includegraphics[width=0.8\linewidth]{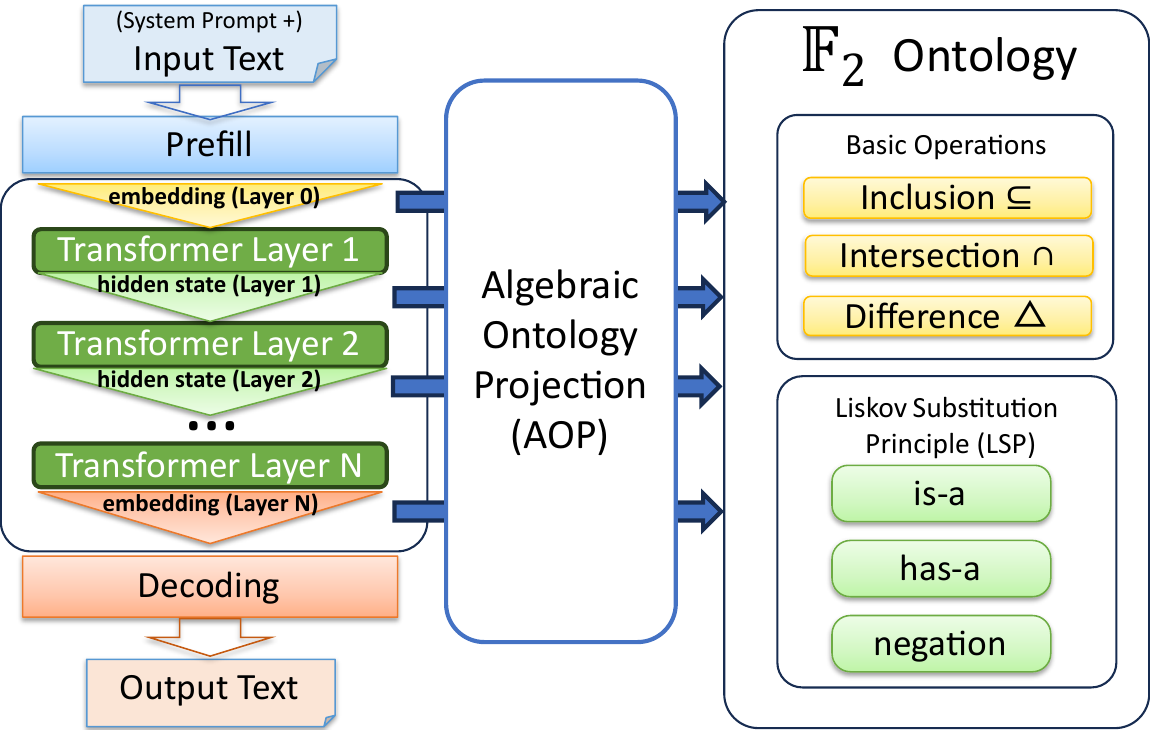}
 \caption{ Overview of Algebraic Ontology 
 Projection~(AOP). 
 Each Transformer layer's hidden states 
 are projected onto an $\mathbb{F}_2$ 
 binary vector space under relational 
 constraints~(is-a, has-a, negation). 
 The system prompt, when present, 
 is processed as prefill and serves 
 as an algebraic boundary condition 
 that configures the hidden state 
 before projection~(Section~\ref{sec:method}).
 }
 \label{fig:aop-overview}
\end{figure}

Concretely, we train a two-layer linear network on just 42
relational pairs drawn from four semantic domains.
Evaluated on held-out entities and relations that never appear
during training, our method achieves up to \textbf{93.33\%
inclusion accuracy} across Gemma-2 and Qwen2.5---
with no model tuning, but through prompt alone---
substantially above chance and competitive
with models trained on orders of magnitude more supervision.
These results hold for concepts entirely absent from the
training set---including cross-domain pairs such as
\textit{Copper}~$\subseteq$~\textit{Metal} and
\textit{Marble}~$\subseteq$~\textit{Rock}---demonstrating
genuine structural generalisation rather than memorisation.

Standard representational analyses---probing classifiers,
Centered Kernel Alignment~(CKA), and Singular Vector
Canonical Correlation Analysis~(SVCCA)---treat meaning as
geometry: concepts are points, and similarity is distance.
This perspective cannot express the asymmetric, transitive
structure of hierarchical relations.
Knowing that \textit{Eagle}~$\subseteq$~\textit{Bird} is not
the same as knowing that Eagle and Bird are nearby in embedding
space: the former is directional and non-destructive, the
latter is symmetric and admits no formal entailment.
Our approach asks a different question: not how close are
these representations, but whether one \emph{contains} the
other in a formal algebraic sense.

We introduce \textbf{Algebraic Ontology Projection~(AOP)},
which maps LLM hidden states into a binary vector
space~$\{0,1\}^n$ under the constraint that hierarchical
relations~(\textit{is-a}) correspond to bitwise inclusion:
\begin{equation}
  \mathbf{a} \odot \mathbf{b} = \mathbf{a}
  \quad \Leftrightarrow \quad A \subseteq B,
  \label{eq:lsp}
\end{equation}
where $\odot$ is elementwise multiplication.
The projection is implemented as a two-layer linear network
with a sharp activation function---structurally analogous to
the language model head~(lm\_head), but targeting a binary
algebraic space rather than a vocabulary distribution.
The projection is learned from a small set of relational
\textbf{algebraic keys}: concept pairs whose relations are
known, used not as statistical training signal but as
structural constraints to unlock latent organisation already
present in the model.
We term the extracted structure an \textbf{Algebraic Ontology}.

To extract hidden states for a given concept, we apply
\textbf{Localized Mean Pooling~(LMP)}: the hidden states
corresponding to the concept's context tokens are averaged
across the relevant layer.
When a system prompt is present, it is processed as a
prefill---shaping the model's internal state---but is
excluded from the averaged representation.
This separation ensures that the projection captures the
concept's representation within the configured context,
rather than a global average over the full input.

A central finding is that relational structure is not uniformly
present across layers.
We introduce \textbf{Semantic Crystallisation~(SC)}, a
dimensionless structural metric defined as:
\begin{equation}
  SC(L) = \bigl(\mu_{rand} - q(L)\bigr) \cdot \sigma_{rand}^2,
\end{equation}
where $q(L) = \mathcal{L}_{alg}(L)/\rho(L)^2$ is the
density-normalised algebraic loss relative to a random
baseline~(Section~\ref{sec:method:sc}).
SC takes values in $[-1, 1]$: positive values indicate
layers where algebraic structure exceeds the random
baseline; $SC < 0$~(\textbf{Semantic Melting}) indicates
active suppression of relational structure.
Crucially, SC scores \emph{predict} zero-shot generalisation
performance across layers, providing a gradient-free criterion
for layer selection without held-out evaluation data.

A second central finding concerns the role of system prompts.
We observe two qualitatively distinct behaviours across model
families.
In \textbf{Autonomous Crystallisation}~(Gemma), unprompted
layers are near-random~($SC \approx 0$), and a structured
system prompt elevates specific layers to $SC > 1.5$.
In \textbf{Induced Crystallisation}~(Qwen), unprompted
conditions yield lower average SC, with specific layers
exhibiting $SC < 0$; structured prompting substantially
reverses this.
These findings reframe the function of system prompts:
beyond behavioural steering, they act as
\textbf{structural configuration signals} that determine
whether---and where---relational organisation emerges in the
model's hidden states.

We further observe a systematic \textbf{Late-layer Collapse}:
in 7 of 10 evaluated conditions, zero-shot accuracy degrades
significantly in the final layers of the model.
Only the combination of using an \emph{instruction-tuned
model}~(Instruct variant) and providing a \emph{structured
system prompt at inference time} maintains both high peak
accuracy and stable performance through the final layer.
These two factors play complementary roles: the
instruction-tuned model provides a stable representational
substrate~(a property of the model weights), while the
system prompt configures the algebraic boundary conditions
that sustain it~(a property of the inference-time input).

Beyond the interpretability findings reported here, the SC
metric opens a direct path toward \emph{quantitative prompt
engineering}: by analysing SC contributions at the token
level, one can identify which tokens in a system prompt
strengthen or weaken relational structure, reducing prompt
design from trial-and-error to a structured optimisation
problem.
We report these results in a companion paper.

Taken together, these results suggest that LLMs do not merely
approximate relational knowledge geometrically: they encode
it in a structured form that is accessible to linear
algebraic projection, raising new possibilities for
interpretable, verifiable, and formally grounded AI systems.

\paragraph{Contributions.}
\begin{enumerate}

  \item \textbf{Algebraic Ontology Projection~(AOP).}
    A two-layer linear projection method that extracts binary
    symbolic representations of \textit{is-a}, \textit{has-a},
    and negation relations from LLM hidden states using a
    minimal set of relational constraints as algebraic keys,
    with demonstrated applicability across mpnet, Qwen2.5, and
    Gemma-2.

  \item \textbf{Semantic Crystallisation~(SC).}
    A quantitative, baseline-calibrated measure of relational
    structure in LLM hidden states that~(i)~reveals two
    distinct modes of structural organisation across model
    families~(Autonomous and Induced Crystallisation),
    and~(ii)~predicts zero-shot generalisation performance,
    enabling gradient-free layer selection.

  \item \textbf{Contextual Structural Configuration.}
    Empirical demonstration that system prompts function as
    structural configuration signals, transitioning hidden
    states between Semantic Melting and Crystallised regimes,
    with the choice of instruction-tuned vs base model and
    the inference-time system prompt playing complementary
    roles in sustaining late-layer stability.

  \item \textbf{Zero-Shot Ontological Generalisation.}
    Up to 93.33\% inclusion accuracy on unseen concept pairs
    from a projection trained on 42 relational constraints,
    with systematic characterisation of Late-layer Collapse
    and the conditions under which relational consistency
    is maintained throughout the model depth.

\end{enumerate}

\section{Related Work}
\label{sec:related}

\paragraph{Probing and representational analysis.}
Linear probing, introduced by Alain \& Bengio~\citep{alain2016understanding}
for analysing intermediate representations in image
classifiers, was subsequently extended to transformer
hidden states.
Hewitt \& Manning~\citep{hewitt2019structural} showed that
syntactic dependency structure can be recovered from BERT
hidden states via a learned linear transformation into a
\emph{distance space}, where proximity encodes syntactic
closeness.
However, distance-based probes are inherently symmetric:
$d(A, B) = d(B, A)$, and cannot represent the directional
structure of ontological relations.
Knowing that representations of \textit{Eagle} and
\textit{Bird} are nearby does not establish whether
Eagle~$\subseteq$~Bird or Bird~$\subseteq$~Eagle.
AOP addresses this by projecting into $\mathbb{F}_2^n$,
where inclusion is directional by construction:
$\mathbf{a} \odot \mathbf{b} = \mathbf{a}$ implies
$A \subseteq B$, not the reverse.
Representational similarity methods including
CKA~\citep{kornblith2019similarity} and
SVCCA~\citep{raghu2017svcca} share the same geometric
limitation, measuring proximity rather than algebraic
structure.

\paragraph{Mechanistic interpretability and monosemanticity.}
A central programme in mechanistic
interpretability~\citep{olah2020zoom} seeks to identify
the computational roles of individual neurons and circuits
within transformer models.
The superposition hypothesis~\citep{elhage2022superposition}
proposes that LLMs encode more features than available
dimensions by exploiting near-orthogonal directions in
activation space.
Anthropic's work on
monosemanticity~\citep{bricken2023monosemanticity,templeton2024scaling}
operationalises this by applying sparse autoencoders~(SAEs)
to decompose superposed representations into interpretable,
monosemantic features---a major advance in identifying
\emph{which features exist} within a model.
AOP is complementary rather than competing: where
monosemanticity research asks \emph{which features are
present}, AOP asks whether those features stand in the
\emph{algebraic relations}---inclusion, intersection,
negation---that formal ontologies require.
Identifying a feature for ``Diamond'' and a feature for
``Mineral'' does not, by itself, establish that the model
represents Diamond~$\subseteq$~Mineral as a formal
algebraic constraint; AOP tests precisely this.

\paragraph{Knowledge representation in LLMs.}
Petroni et al.~\citep{petroni2019language} demonstrated
that LLMs encode factual and relational knowledge without
fine-tuning, evidenced by strong performance on cloze-style
queries evaluated at the output level.
This behavioural finding raises the deeper question of
whether the underlying representations satisfy the formal
algebraic constraints that define those relations.
AOP provides a structural counterpart: not only can LLMs
answer relational queries correctly, but their internal
representations satisfy the formal algebraic constraints
that define those relations---a stronger claim that goes
beyond output accuracy to the representational substrate.

Meng et al.~\citep{meng2022locating} advanced this by
using causal tracing to localise factual associations
within specific MLP layers---analogous to identifying
definition points in a dataflow analysis.
Their results confirm an important role for mid-layer
feed-forward modules in storing factual associations,
but the side effects of editing these associations on
neighbouring knowledge remain difficult to control,
a consequence of the non-linear information mixing
across layers.
AOP takes a complementary approach: rather than
intervening in the forward pass, it identifies layers
where algebraic structure is already accessible,
without perturbation.

Burns et al.~\citep{burns2022discovering} introduced
Contrast-Consistent Search~(CCS), which recovers scalar
truth values from hidden states without supervision by
searching for a direction that separates true from false
propositions---the approach most closely related in spirit
to AOP.
CCS applies a one-dimensional linear projection
$\sigma(\boldsymbol{\theta}^\top \mathbf{x} + b)$,
which can be viewed as a scalar special case of AOP's
$n$-dimensional $\mathbb{F}_2$ projection.
The zero-shot baselines in their evaluation operate at
the output level via next-token log probabilities rather
than hidden state geometry, with honest zero-shot
accuracy remaining below 80\% on most benchmarks.
AOP differs from CCS in two respects: it recovers a full
algebraic system---closed under inclusion, intersection,
and negation---rather than a scalar truth value, and it
achieves up to 93.33\% zero-shot accuracy on unseen
concept pairs trained on a limited set of relational
constraints.

\paragraph{Formal methods and neural networks.}
Nanda et al.~\citep{nanda2023progress} demonstrated that
small transformers solving modular arithmetic develop
internal Fourier representations corresponding to the
periodic structure of $\mathbb{Z}/p\mathbb{Z}$, and
undergo a three-stage learning process---memorization,
circuit formation, and cleanup---in which algebraic
structure emerges spontaneously without explicit
supervision.
The authors note that their progress measures are specific
to small networks on a single algorithmic task, and that
task-independent measures are necessary for broader
applicability.
AOP addresses this by grounding the projection in
$\mathbb{F}_2$---a structure whose algebraic closure
applies regardless of task---and demonstrates consistent
results across multiple model families and semantic
domains without task-specific training.

Geiger et al.~\citep{geiger2021causal} developed causal
abstraction to test whether neural networks implement
specific task-defined causal models via interchange
interventions.
AOP adopts a different computational
model---algebraic relations over $\mathbb{F}_2$ grounded
in the Liskov Substitution Principle---and asks whether
pre-trained hidden states satisfy this structure, rather
than verifying a task-specific causal hypothesis.

Neuro-symbolic approaches~\citep{garcez2022neural} have
sought to integrate logical constraints with neural
computation, typically by training models on logical
supervision or enforcing constraints at the output level.
AOP differs in that it does not train LLMs to satisfy
logical constraints: it \emph{projects} pre-trained
representations to test whether constraints are already
latently satisfied.
The distinction is between teaching a model logic and
discovering whether it has already acquired it.

\noindent
Collectively, existing methods analyse LLM representations
geometrically, behaviourally, through causal intervention,
or through scalar feature decomposition---but none
directly measures whether hidden states satisfy the
algebraic invariants required by formal ontologies as a
closed system.
AOP addresses this gap, and in doing so reveals systematic
layer-dependent patterns---Semantic Crystallisation and
Late-layer Collapse---that are invisible to existing
analyses.

\section{Algebraic Ontology Projection}
\label{sec:method}

We introduce \textbf{Algebraic Ontology Projection~(AOP)}, a
framework that extracts formal relational structure from LLM
hidden states by projecting them into a binary vector space
equipped with algebraic operations.

\subsection{Binary Algebraic Structure over
  \texorpdfstring{$\mathbb{F}_2$}{F2}}
\label{sec:method:f2}

\paragraph{Representing concepts as binary vectors.}
We represent each concept as a binary vector
$\mathbf{a} \in \{0,1\}^n$, where each bit encodes the
presence or absence of a latent semantic feature.
The key operation in $\mathbb{F}_2^n$ is the elementwise
product~$\odot$~(bitwise AND), which computes the
\emph{intersection} of two concept representations:
$\mathbf{a} \odot \mathbf{b}$ extracts the features shared
by both $A$ and $B$.

Under this interpretation, the \textit{is-a} relation
$A \subseteq B$---``every $A$ is a $B$''---is defined as:
\begin{equation}
  A \subseteq B
  \;\Leftrightarrow\;
  \mathbf{a} \odot \mathbf{b} = \mathbf{a},
  \label{eq:isa}
\end{equation}
meaning that the features of $A$ are entirely contained
within those of $B$: the intersection of $A$ and $B$
recovers $A$ itself.
Note carefully that this is a condition on
\emph{intersection}, not on bit count:
$\mathbf{a} \odot \mathbf{b} = \mathbf{a}$ requires that
every bit active in $\mathbf{a}$ is also active in
$\mathbf{b}$, but $\mathbf{b}$ may activate additional bits.

\paragraph{Feature accumulation across the hierarchy.}
A crucial consequence of this formulation is that
\emph{subordinate concepts accumulate more active bits than
superordinate concepts}.
Consider the chain
\textit{Animal}~$\supseteq$~\textit{Insect}~$\supseteq$%
~\textit{Beetle}~$\supseteq$~\textit{StagBeetle}:
\textit{StagBeetle} inherits all features of
\textit{Beetle}, which inherits all features of
\textit{Insect}, which inherits all features of
\textit{Animal}---plus, at each level, concept-specific
features are added.
\textit{StagBeetle} therefore activates strictly more bits
than \textit{Animal}, not fewer.
This mirrors the intuition that more specific concepts are
\emph{richer in features} than abstract ones, and is
reinforced by LSP inheritance: \textit{has-a} attributes
are propagated downward, further enriching subordinate
representations.

\paragraph{Why binary?}
The binary vector space $\{0,1\}^n$ is isomorphic to the
power set lattice $2^{[n]}$---the natural algebraic
structure of ontological hierarchies, in which concepts
are sets of features and containment corresponds to
subsumption.
Formally, $\{0,1\}^n$ equipped with elementwise addition
modulo~2 and elementwise multiplication~(AND) forms the
Galois Field $\mathbb{F}_2^n$, guaranteeing \emph{algebraic
closure}: any composition of bitwise operations remains
within $\mathbb{F}_2^n$ without requiring normalisation
such as softmax.

\paragraph{Relational operations.}
The full set of ontological relations maps directly to
$\mathbb{F}_2^n$ operations:
\begin{align}
  A \subseteq B
    &\;\Leftrightarrow\;
      \mathbf{a} \odot \mathbf{b} = \mathbf{a}
    \quad\text{(\textit{is-a})}
    \label{eq:rel_isa} \\
  A \cap B
    &\;\Leftrightarrow\;
      \mathbf{a} \odot \mathbf{b}
    \quad\text{(feature intersection)}
    \label{eq:rel_intersection} \\
  A \triangle B
    &\;\Leftrightarrow\;
      \mathbf{a} \oplus \mathbf{b}
    \quad\text{(symmetric difference)}
    \label{eq:rel_xor}
\end{align}
The \textit{has-a} relation~($A$ has part $P$) is defined
analogously: $\mathbf{p} \odot \mathbf{a} = \mathbf{p}$,
i.e., all features of the part are present in the whole.
Negation~($A$ is not $B$) requires
$\mathbf{a} \odot \mathbf{b} = \mathbf{0}$: no shared
features.
We term the structure defined by these operations an
\textbf{Algebraic Ontology}.

\subsection{Projection Architecture}
\label{sec:method:arch}

The AOP maps a hidden state
$\mathbf{h} \in \mathbb{R}^d$ to a soft-binary vector
$\mathbf{z} \in [0,1]^n$ via a two-stage transformation:
\begin{equation}
  \mathbf{z} = \sigma\!\left(
    \gamma \cdot W_2 \cdot
    \tanh(W_1 \mathbf{h} - \boldsymbol{\theta})
  \right),
  \label{eq:aop}
\end{equation}
where $W_1 \in \mathbb{R}^{n \times d}$ is the
\textbf{attribute extraction layer};
$\boldsymbol{\theta} \in \mathbb{R}^n$ is a
\textbf{learnable per-dimension threshold};
$W_2 \in \mathbb{R}^{n \times n}$ is the
\textbf{logical mapping layer};
$\gamma = 4.0$ is a fixed sharpness parameter; and
$\sigma$ is the sigmoid function applied elementwise.

\paragraph{Role of the adaptive threshold.}
The learnable threshold $\boldsymbol{\theta}$ independently
calibrates each bit's activation tendency.
Dimensions with $\theta_i < 0$ activate readily,
corresponding to abstract features shared by many concepts
(e.g., features common to all \textit{Animal} instances).
Dimensions with $\theta_i > 0$ activate only under strong
evidence, corresponding to specific features held by few
concepts (e.g., features unique to \textit{StagBeetle}).
This asymmetry reflects the feature accumulation property
of Section~\ref{sec:method:f2}: superordinate concepts
activate few, broadly-shared features; subordinate concepts
activate many, including all inherited features.

\paragraph{Binarisation.}
During training, outputs are soft-valued in $[0,1]$,
enabling gradient-based optimisation.
During evaluation, outputs are binarised at $0.5$.
The sharpness parameter $\gamma = 4.0$ encourages outputs
to concentrate near $\{0,1\}$ during training.

\paragraph{Analogy with lm\_head.}
AOP is structurally analogous to the standard language model
head: both apply a linear transformation to LLM hidden
states to decode structured information.
The key differences are the target space~($\mathbb{F}_2^n$
rather than a vocabulary distribution) and the
discretisation mechanism~(adaptive thresholding rather than
softmax).
Where lm\_head must impose closure over a probability
simplex via softmax, AOP inherits algebraic closure from
the field structure of $\mathbb{F}_2^n$ by construction.

\subsection{Training Protocol}
\label{sec:method:training}

\paragraph{Algebraic keys.}
AOP is trained on a minimal set of relational pairs that we
term \textbf{algebraic keys}: formal constraints whose
relational type is known, used as algebraic anchors that
render the model's latent structure legible rather than as
statistical training data.

\paragraph{Dataset structure.}
The relational dataset is organised into four progressive
stages~(levels 1, 2, 4, and 8), spanning four semantic
domains~(biological, mineral, physical, and abstract),
each introducing new concepts and relations to expand
both the semantic diversity and the difficulty of the
algebraic constraints.
All four stages are used simultaneously during training.
Level~1 establishes a basic insect--animal hierarchy;
levels~2 and~4 extend this with additional biological
concepts; level~8 introduces a heterogeneous mineral
domain, testing cross-domain generalisation of the
projection.
The complete dataset comprises 42~training pairs~(15
\textit{is-a}, 12~\textit{has-a}, 15~negation) and
13~independent evaluation pairs~(i\_neg,
$D_{val} \cap D_{train} = \emptyset$).
The independent evaluation set consists of concept pairs
drawn from outside the semantic domains of the training
data---pairing abstract concepts with physical
attributes~(e.g., \textit{Idea}~vs \textit{Legs})
and cross-domain entities~(e.g., \textit{DNA}~vs
\textit{Cloud}).
These pairs are never used during training and serve
exclusively to verify structural generalisation.
Full dataset details are provided in
Appendix~\ref{app:dataset}.

\paragraph{Loss function.}
The training objective jointly enforces four classes of
algebraic constraint:

\textbf{(i) Is-a inclusion~($\mathcal{L}_{isa}$):}
penalises cases where the parent representation lacks bits
that the child representation activates, enforcing
$\mathbf{a}_{child} \odot \mathbf{a}_{parent} =
\mathbf{a}_{child}$.

\textbf{(ii) Has-a inclusion~($\mathcal{L}_{has}$):}
penalises cases where the whole concept lacks bits present
in the part representation, enforcing
$\mathbf{a}_{part} \odot \mathbf{a}_{parent} =
\mathbf{a}_{part}$.
This term is weighted more strongly than $\mathcal{L}_{isa}$,
reflecting the tighter compositional constraint of
part-whole relations.

\textbf{(iii) LSP inheritance~($\mathcal{L}_{lsp}$):}
the most critical constraint, enforcing that the child
concept inherits the intersection of parent and part
features:
\begin{equation}
  \mathcal{L}_{lsp} =
    \mathbb{E}\!\left[
      \max\!\left(
        \mathbf{a}_{parent} \odot \mathbf{a}_{part}
        - \mathbf{a}_{child},\;
        \mathbf{0}
      \right)
    \right],
  \label{eq:lsp_loss}
\end{equation}
where $\odot$ extracts the features that the parent
possesses \emph{as part}, which the child must also
possess by the Liskov Substitution Principle.

\textbf{(iv) Separation and density regularisation:}
three additional terms prevent degenerate solutions.
A \emph{separation} term enforces that negation pairs
maintain an appropriate Hamming distance, discouraging both
complete overlap and complete orthogonality.
A \emph{density} term targets different bit densities for
superordinate and subordinate concepts, reflecting the
feature accumulation property: superordinate concepts
should activate fewer bits~(more abstract) and subordinate
concepts more bits~(more specific).
An \emph{anti-zero} term penalises representations in which
no bits are active, preventing the trivially satisfied but
uninformative zero solution.
An \emph{orthogonality} term prevents concept-specific bits
of the child from overlapping with part features inherited
from the parent, preserving the distinctiveness of each
level of the hierarchy.

\paragraph{Zero-shot evaluation metric.}
For evaluation, we measure the \textbf{inclusion score}:
\begin{equation}
  \text{Inclusion}(A, B) =
    \frac{|\mathbf{a} \odot \mathbf{b}|}{|\mathbf{b}|},
  \label{eq:inclusion}
\end{equation}
where $|\cdot|$ denotes the Hamming weight.
This measures the fraction of $B$'s active bits that are
also active in $A$---directly operationalising the
$\mathbb{F}_2$ definition of \textit{is-a}:
if $A \subseteq B$, then $B$'s bits are a subset of $A$'s bits,
so a higher-level concept~($B$) activates fewer bits
than its subordinate~($A$).
A pair is classified as \textit{is-a} if
$\text{Inclusion}(A,B) \geq \tau = 0.7$ and the
complementary Hamming distance satisfies $\delta = 0.1$.

\subsection{Semantic Crystallisation~(SC)}
\label{sec:method:sc}

\paragraph{Motivation.}
The degree to which a given layer's hidden states satisfy
$\mathbb{F}_2$ algebraic constraints varies substantially
across layers and model configurations.
To enable dimensionless, cross-architecture comparison,
we introduce \textbf{Semantic Crystallisation~(SC)}: a
dimensionless indicator that quantifies the emergence of
algebraic order within the latent representation space,
evaluated relative to the intrinsic variance of the
model's stochastic baseline.

\paragraph{Density-invariant algebraic loss.}
Let $\mathcal{L}_{alg}$ denote the raw algebraic
inconsistency loss density and $\rho$ the mean bit activation rate
(\emph{bit density}) of the projected representations.
We define the \textbf{density-normalised algebraic loss}:
\begin{equation}
  q(L) = \frac{\mathcal{L}_{alg}(L)}{\rho(L)^2},
  \label{eq:q}
\end{equation}
where the $1/\rho^2$ factor isolates the structural
alignment from the quadratic increase in stochastic
collisions inherent in high-density representations.
In practice, $\rho(L)$ is estimated from the \textit{is-a}
and \textit{has-a} training pairs only; negation pairs are
excluded from this estimate because they systematically
activate fewer bits, which would introduce a downward bias
into the density approximation and adversely affect SC
values.

\paragraph{Stochastic baseline and scaling factor.}
For each model architecture, we apply AOP to a randomly
initialised model of identical architecture and compute:
\begin{equation}
  \mu_{rand} = \mathbb{E}[q_{rand}], \qquad
  \sigma_{rand}^2 = \mathrm{Var}(q_{rand}),
  \label{eq:baseline}
\end{equation}
where $\mu_{rand}$ and $\sigma_{rand}^2$ characterise
the expected noise floor and the intrinsic fluctuations
of the architecture, respectively.

\paragraph{SC as a dimensionless order parameter.}
\begin{equation}
  SC(L) = \bigl(\mu_{rand} - q(L)\bigr) \cdot \sigma_{rand}^2,
  \label{eq:sc}
\end{equation}
$SC$ is a \textbf{dimensionless quantity}.
The term $(\mu_{rand} - q(L))$ represents the degree to
which algebraic structure has emerged from the stochastic
background, while $\sigma_{rand}^2$ acts as a
characteristic scaling factor that ensures crystallisation
is evaluated relative to the architecture's inherent
capacity for representational diversity, enabling
consistent comparison across heterogeneous model families.
SC typically takes values between $[-1, 1]$.

We distinguish three regimes:
\begin{itemize}
  \item \textbf{Crystalline phase}~($SC \approx 1$):
    logical structures are dominant and robust;
    relational constraints are satisfied well above chance.
  \item \textbf{Gas/Vacuum phase}~($SC \approx 0$):
    maximum entropy state; representations are
    indistinguishable from the random baseline.
  \item \textbf{Collapsed phase}~($SC < 0$):
    algebraic structures are actively suppressed below
    the random baseline~(\textbf{Semantic Melting}).
\end{itemize}

\paragraph{SC as a layer selection criterion.}
We demonstrate empirically~(Section~\ref{sec:experiments:sc})
that SC scores predict zero-shot generalisation performance,
providing a gradient-free criterion for selecting the
optimal projection layer without held-out evaluation data.

\subsection{Localized Mean Pooling~(LMP)}
\label{sec:method:lmp}

\paragraph{Concept representation.}
To obtain the hidden state representation of a concept, we
query the model with a short context string and average the
hidden states of the context tokens at the target layer:
\begin{equation}
  \mathbf{h}_{concept} =
    \frac{1}{|T_{ctx}|}
    \sum_{t \in T_{ctx}} \mathbf{h}_t^{(L)},
  \label{eq:lmp}
\end{equation}
where $T_{ctx}$ indexes the tokens of the concept context
and $L$ is the target layer.

\paragraph{System prompt separation.}
When a system prompt is present, it is processed as
\emph{prefill}---configuring the model's hidden state as
an algebraic boundary condition---but its tokens are
excluded from the average in Equation~\ref{eq:lmp}:
\begin{equation}
  \mathbf{h}_{concept} =
  \operatorname{LMP}\!\left(
    \underbrace{[\text{system prompt}]}_{\text{prefill only}}
    +
    \underbrace{[\text{concept context}]}_{\text{averaged}}
  \right).
  \label{eq:lmp_sep}
\end{equation}
This separation reflects the distinct roles of global
context as an algebraic boundary condition and local
context as the concept query.

\paragraph{Scope and limitations.}
LMP is effective when the concept context is short and the
target concept dominates the token sequence.
For longer contexts, mean pooling dilutes the concept
signal, degrading projection quality.
All experiments use minimal context strings to satisfy
this condition.
Extension to token-level projection is a direction for
future work.

\section{Experiments}
\label{sec:experiments}

\subsection{Experimental Setup}
\label{sec:experiments:setup}

\paragraph{Models.}
We evaluate AOP on three model families:
Gemma-2~\citep{gemma2024}~(2B parameters,
26~Transformer layers, layer indices 0--26,
hidden size~2048),
Qwen2.5~\citep{qwen2024}~(1.5B parameters,
28~Transformer layers, layer indices 0--28,
hidden size~1536), and
mpnet~\citep{reimers2019sentence}~(12~Transformer layers,
layer indices 0--12, hidden size~768),
where layer index~0 denotes the input embedding layer
and the final index denotes the output embedding layer.
All models are loaded in bfloat16 precision.
We note that Gemma-2 and Qwen2.5 differ in parameter
count and architecture; differences in AOP performance
across model families should be interpreted accordingly.
We evaluate both instruction-tuned and base variants
where available.
For each architecture, a randomly initialised model of
identical configuration serves as the algebraic noise
baseline for SC computation~(Section~\ref{sec:method:sc}).
All models are evaluated without fine-tuning; only the AOP
projection weights are trained.

\paragraph{Projection dimension.}
We use a fixed projection dimension of $n = 2048$ across
all models, matching the hidden state dimension of Gemma
and providing a common basis for cross-model comparison.
The effect of projection dimension on AOP performance is
left for future investigation.

\paragraph{Training dataset.}
The algebraic key dataset comprises 42~symbolic constraints
across four semantic domains, as described in
Section~\ref{sec:method:training}.
A held-out evaluation set of 15~concept pairs~(unseen
entities and relations) is used for zero-shot
generalisation assessment.

\paragraph{Optimisation.}
We optimise using AdamW with weight decay $10^{-2}$ and a
ReduceLROnPlateau scheduler~(reduction factor $0.5$,
patience $200$ steps).
Loss terms employ SoftPlus activations in place of ReLU
for numerical stability across model architectures of
varying hidden state dimensions.
Training is terminated upon detection of loss
instability~(buckling), and the checkpoint achieving the
lowest training loss prior to instability is retained for
evaluation.

\paragraph{System prompt conditions.}
We evaluate each model under two conditions:
\textbf{no-prompt}~(minimal prefill token only) and
\textbf{optimized prompt}~(structured system prompt
processed as prefill).
The optimized prompt used in all experiments is:

\begin{quote}
\texttt{You are an expert tax formal hierarchy.
Constraints: Focus on `is-a' and `has-a'.
Suppress colloquial or metaphorical.}
\end{quote}

This prompt instructs the model to interpret input
terms exclusively within a formal taxonomic hierarchy,
suppressing colloquial or metaphorical extensions.
The prompt was designed to be minimal while maximising
Semantic Crystallisation~(SC) scores, and is held
constant across all models and conditions.

\paragraph{Evaluation metrics.}
We report~(i)~\textbf{SC scores} per layer, computed
against the random baseline~(Equation~\ref{eq:sc}),
and~(ii)~\textbf{zero-shot inclusion accuracy} on
held-out concept pairs~(Equation~\ref{eq:inclusion}),
with thresholds $\tau = 0.7$ and $\delta = 0.1$.

\subsection{Layer-wise Semantic Crystallisation}
\label{sec:experiments:sc}

\begin{figure}[t]
  \centering
  \includegraphics[width=\linewidth]{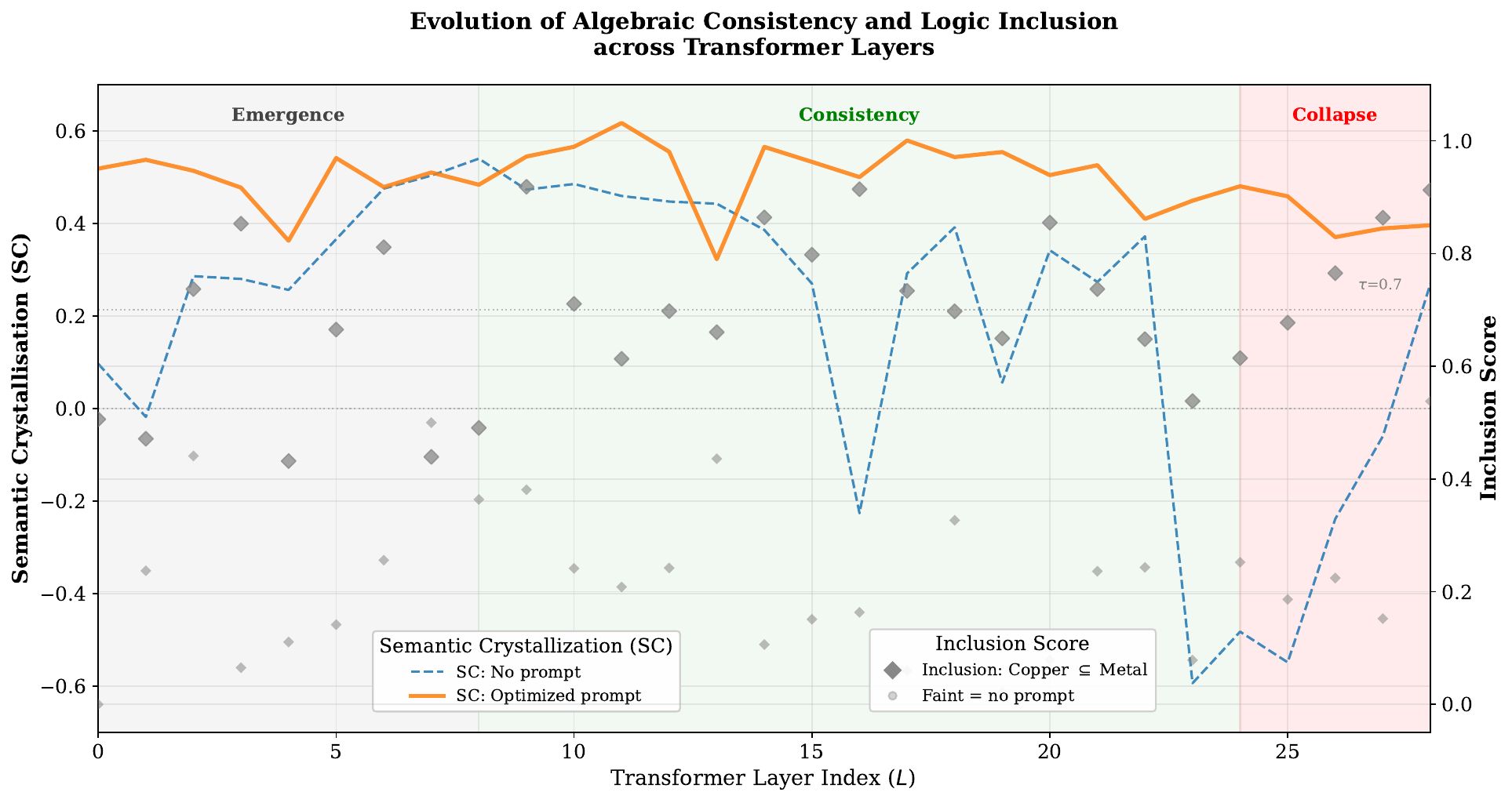}
  \caption{
    Semantic Crystallisation~(SC) scores~(left axis) and
    zero-shot inclusion scores~(right axis) across
    Transformer layer indices for Qwen2.5, under
    no-prompt~(dashed) and optimized~(solid) conditions.
    The horizontal line at $SC = 0$ represents the random
    baseline.
    Three regimes are annotated: \textbf{Emergence}~(early
    layers), \textbf{Consistency}~(middle layers), and
    \textbf{Collapse}~(final layers).
    Coloured dots indicate zero-shot inclusion scores
    ~(Equation~\ref{eq:inclusion}) at each layer~(right axis),
    demonstrating the correspondence between SC and
    zero-shot generalisation performance.
  }
  \label{fig:sc_layers}
\end{figure}

\paragraph{Three regimes of algebraic order.}
Figure~\ref{fig:sc_layers} shows SC scores across all
layers for Gemma and Qwen under no-prompt and optimized
conditions, relative to the random baseline~($SC = 0$).
We observe three qualitatively distinct regimes:

\textbf{Emergence}~(early layers): SC transitions from
near-zero or negative values toward positive values as the
model begins processing the input.
The rate and extent of this emergence differs markedly
across models and conditions.

\textbf{Consistency}~(middle layers): SC stabilises at
positive values, indicating sustained algebraic order.
This regime is most pronounced under the optimized
condition and in instruction-tuned models.

\textbf{Collapse}~(final layers): SC degrades in the
majority of conditions, consistent with the Late-layer
Collapse phenomenon described in
Section~\ref{sec:experiments:collapse}.

\paragraph{Two crystallisation modes.}
We observe two qualitatively distinct patterns across
model families.

In \textbf{Autonomous Crystallisation}~(Gemma), the
no-prompt condition yields $SC \approx 0$ across most
layers~(average $SC = -0.104$), with the highest SC
observed at the input embedding layer~(layer index~0),
while the optimized prompt condition
elevates specific layers substantially~(average
$SC = +0.093$, best layer $SC = 0.148$ at layer index~14).
This suggests that $\mathbb{F}_2$ structure is latent in
Gemma but requires a structured context to become
extractable in deeper layers.

In \textbf{Induced Crystallisation}~(Qwen, mpnet), the no-prompt
condition yields modest average SC across layers~(average
$SC = +0.193$ for Qwen), with specific layers exhibiting negative
values.
The optimized prompt condition substantially elevates
algebraic consistency~(average $SC = +0.493$,
best layer $SC = 0.617$ at layer index~11),
representing a substantial transition between representational
regimes.
This effect is most pronounced in mpnet: the no-prompt
condition yields Avg~$SC = -0.174$, while the optimized
prompt elevates this to $+0.266$---the most dramatic
Induced Crystallisation observed across all evaluated
conditions.

\paragraph{SC predicts zero-shot performance.}
A key practical finding is that SC scores predict zero-shot
generalisation accuracy across layers: the layer achieving
the highest SC consistently yields the highest zero-shot
inclusion accuracy~(Section~\ref{sec:experiments:zst}).
This provides a gradient-free criterion for layer selection
without held-out evaluation data.

\subsection{Zero-Shot Ontological Generalisation}
\label{sec:experiments:zst}

\paragraph{Task.}
We evaluate AOP's ability to recover ontological relations
for concept pairs that never appear in the 42-pair training
set.
The evaluation set~($D_{ZST}$) comprises 15~pairs spanning
9~positive \textit{is-a} relations and 6~negative relations,
covering biological and mineral domains
~(full list in Appendix~\ref{app:zst}).
All concepts in $D_{ZST}$ are drawn from outside the
training vocabulary, ensuring
$D_{ZST} \cap D_{train} = \emptyset$.

\begin{figure}[t]
  \centering
  \includegraphics[width=\linewidth]{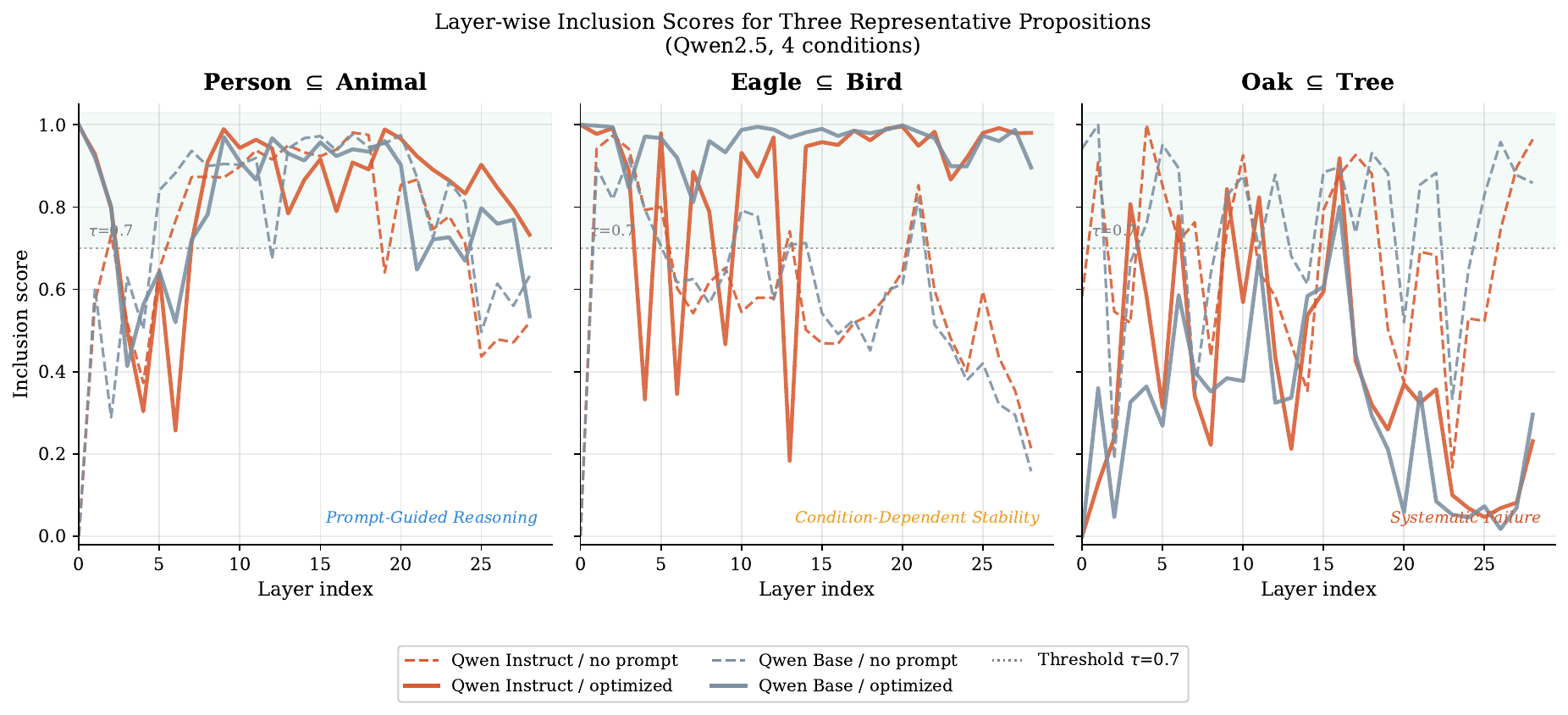}
  \caption{
    Layer-wise inclusion scores~(Equation~\ref{eq:inclusion})
    for three representative propositions across four Qwen2.5
    conditions.
    \textbf{Left}~(\textit{Person}~$\subseteq$~\textit{Animal},
    \textit{Prompt-Guided Reasoning}):
    without a structured prompt, the model withholds commitment
    to the biological classification of persons as animals---
    a philosophically contested assertion---and inclusion scores
    remain unstable across layers.
    The optimized prompt resolves this ambiguity, sustaining
    high scores throughout.
    \textbf{Centre}~(\textit{Eagle}~$\subseteq$~\textit{Bird},
    \textit{Condition-Dependent Stability}):
    instruction-tuned models with optimized prompt maintain
    high inclusion scores throughout, while base models
    and no-prompt conditions exhibit Late-layer Collapse
    beyond layer~20.
    \textbf{Right}~(\textit{Oak}~$\subseteq$~\textit{Tree},
    \textit{Systematic Failure}):
    inclusion scores collapse across all conditions,
    attributable to the lexical ambiguity of ``Oak''
    in the model's training distribution~(Section~\ref{sec:discussion:limitations}).
    The horizontal dotted line indicates the inclusion
    threshold $\tau = 0.7$.
    Warm colours~(red): instruction-tuned models.
    Cool colours~(grey-blue): base models.
    Solid lines: optimized prompt.
    Dashed lines: no-prompt condition.
  }
  \label{fig:zst_layers}
\end{figure}

\paragraph{Results.}
Table~\ref{tab:zst} reports zero-shot accuracy at the best
SC layer for each model-condition combination.
Figure~\ref{fig:zst_layers} illustrates three qualitatively
distinct patterns observed across propositions:
prompt-guided reasoning~(\textit{Person}~$\subseteq$~\textit{Animal}),
condition-dependent stability~(\textit{Eagle}~$\subseteq$~\textit{Bird}),
and systematic failure due to lexical ambiguity~(\textit{Oak}~$\subseteq$~\textit{Tree}).

\begin{table}[t]
  \centering
  \caption{
    Zero-shot ontological generalisation accuracy~(\%)
    at the best SC layer for each model and condition.
    ``Best layer'' denotes the layer index achieving
    the highest SC score.
    Overall accuracy, inclusion accuracy, and Hamming
    accuracy are reported with thresholds
    $\tau = 0.7$, $\delta = 0.1$.
  }
  \label{tab:zst}
  \begin{tabular}{llccccc}
    \toprule
    Model & Condition & Best Layer & Max SC
          & Overall & Inclusion & Hamming \\
    \midrule
    Gemma-2 (Instruct) & optimized  & 14 & 0.148 & 93.33 & 93.33 & 93.33 \\
    Gemma-2 (Instruct) & no-prompt  &  0 & 0.122 & 53.33 & 60.00 & 80.00 \\
    Gemma-2 (Base)     & optimized  &  5 & 0.172 & 53.33 & 60.00 & 60.00 \\
    Gemma-2 (Base)     & no-prompt  &  2 & 0.086 & 66.67 & 66.67 & 86.67 \\
    \midrule
    Qwen2.5 (Instruct) & optimized  & 11 & 0.617 & 86.67 & 86.67 & 93.33 \\
    Qwen2.5 (Instruct) & no-prompt  &  8 & 0.540 & 53.33 & 60.00 & 60.00 \\
    Qwen2.5 (Base)     & optimized  & 17 & 0.294 & 73.33 & 73.33 & 80.00 \\
    Qwen2.5 (Base)     & no-prompt  &  7 & 0.269 & 53.33 & 73.33 & 66.67 \\
    \midrule
    mpnet              & optimized  &  3 & 0.465 & 86.67 & 86.67 & 86.67 \\
    mpnet              & no-prompt  &  3 & 0.449 & 60.00 & 60.00 & 60.00 \\
    \bottomrule
  \end{tabular}
\end{table}

The optimized prompt condition on Gemma-2~(Instruct)
achieves the highest accuracy: \textbf{93.33\%} overall,
inclusion, and Hamming accuracy~(layer index~14,
$SC = 0.148$).
mpnet with optimized prompt achieves 86.67\% overall
accuracy~(layer index~3, $SC = 0.465$).
Qwen2.5~(Instruct) with optimized prompt achieves
86.67\%~(layer index~11, $SC = 0.617$).
The no-prompt condition yields substantially lower
performance across all models, consistent with the lower
SC values observed in
Section~\ref{sec:experiments:sc}.

\paragraph{Systematic failure: plant domain.}
The pair \textit{Oak}~$\subseteq$~\textit{Tree} fails
across all models and conditions at the final layer,
and no condition achieves stable accuracy throughout.
Partial success is observed in specific layers for all
models~(early layers for Gemma-2, intermediate layers
for Qwen2.5 and mpnet), but inclusion scores are
highly unstable and do not meet the threshold consistently.
We attribute this to the lexical ambiguity of ``Oak''
in the model's training distribution, where it appears
frequently as a proper noun~(place name, surname)
unrelated to botanical taxonomy.
This illustrates a principled limitation of AOP under
concept ambiguity: when mean pooling mixes multiple
semantic senses, the resulting representation satisfies
no single ontological constraint cleanly.

\paragraph{Relationship between SC and accuracy.}
The correspondence between peak SC layer and peak accuracy
layer holds consistently across models, validating SC as
a practical layer selection criterion.
In its current formulation, however, SC is a
\textbf{necessary but not sufficient} condition for high
zero-shot accuracy: a strongly negative SC reliably
predicts degraded accuracy~(insulation failure), but a
high SC does not guarantee high accuracy, as SC currently
captures only the insulation dimension of algebraic
consistency~(Section~\ref{sec:discussion:sc}).

\subsection{Late-layer Collapse}
\label{sec:experiments:collapse}

We observe a systematic degradation of zero-shot accuracy
in the final layers of the model, which we term
\textbf{Late-layer Collapse}.
In 7~of~10~evaluated conditions, the accuracy at the final
layer falls more than 10~percentage points below peak
accuracy.

\begin{table}[t]
  \centering
  \caption{
    Late-layer Collapse summary.
    Peak accuracy, final-layer accuracy, and end-layer
    stability~(average inclusion score over final 5 layers)
    are reported for each model-condition combination.
    $\dagger$~indicates conditions that maintain both high
    peak accuracy and late-layer stability.
  }
  \label{tab:collapse}
  \begin{tabular}{llcccc}
    \toprule
    Model & Condition
          & Peak~(\%) & Peak Layer
          & Final~(\%) & Stable \\
    \midrule
    Gemma-2 (Instruct) & optimized  & 93.33 & 14 & 93.33 & $\dagger$ \\
    Gemma-2 (Instruct) & no-prompt  & 53.33 &  0 & 53.33 &           \\
    Gemma-2 (Base)     & optimized  & 53.33 &  5 & 53.33 &           \\
    Gemma-2 (Base)     & no-prompt  & 66.67 &  2 & 66.67 &           \\
    \midrule
    Qwen2.5 (Instruct) & optimized  & 86.67 & 11 & 73.33 & $\dagger$ \\
    Qwen2.5 (Instruct) & no-prompt  & 53.33 &  8 & 46.67 &           \\
    Qwen2.5 (Base)     & optimized  & 73.33 & 17 & 60.00 &           \\
    Qwen2.5 (Base)     & no-prompt  & 53.33 &  7 & 40.00 &           \\
    \midrule
    mpnet              & optimized  & 86.67 &  3 & 60.00 & $\dagger$ \\
    mpnet              & no-prompt  & 60.00 &  3 & 60.00 &           \\
    \bottomrule
  \end{tabular}
\end{table}

\paragraph{Stability conditions.}
Among evaluated conditions, \textbf{optimized/Gemma-2~(Instruct)}
achieves the highest peak accuracy~(93.33\%) while maintaining
the most stable late-layer performance.
Base model variants and no-prompt conditions uniformly
exhibit greater collapse regardless of peak accuracy,
consistent with the complementary roles of instruction
tuning and structured prompting.

\paragraph{Complementary roles.}
Under the conditions evaluated here, instruction tuning
and system prompting appear to play complementary roles:
the instruction-tuned model provides a more stable
representational substrate, while the system prompt
configures the algebraic boundary conditions that
determine whether $\mathbb{F}_2$ structure is accessible.
Whether a carefully designed prompt can substitute for
instruction tuning in base models---achieving comparable
stability through prompt design alone---remains an open
question for future work.

\paragraph{Logic Cliff.}
In some conditions, we observe an abrupt accuracy drop
exceeding 25~percentage points within a single layer
transition, which we term a \textbf{Logic Cliff}.
Logic Cliffs are observed at specific intermediate layers
across multiple conditions, most prominently in
optimized/Gemma-2~(layer index~6) and
optimized/Qwen2.5~(layer index~5), suggesting critical transitions
in representational mode that can disrupt previously
established algebraic structure.

\subsection{Base versus Instruction-tuned Models}
\label{sec:experiments:base_instruct}

\begin{table}[t]
  \centering
  \caption{
    Comparison of SC profiles for base and
    instruction-tuned model variants.
    Average SC~(Avg~SC) and maximum SC~(Max~SC) are
    reported for no-prompt and optimized conditions.
    Spark Prompt results are included for base variants.
  }
  \label{tab:base_instruct}
  \begin{tabular}{llcccc}
    \toprule
    Model & Condition & Best Layer & Max SC & Avg SC \\
    \midrule
    Gemma-2 (Instruct) & no-prompt  &  0 & 0.122 & $-0.104$ \\
    Gemma-2 (Instruct) & optimized  & 14 & 0.148 & $+0.093$ \\
    Gemma-2 (Base)     & no-prompt  &  2 & 0.086 & $-0.216$ \\
    Gemma-2 (Base)     & optimized  &  5 & 0.172 & $+0.047$ \\
    \midrule
    Qwen2.5 (Instruct) & no-prompt  &  8 & 0.540 & $+0.193$ \\
    Qwen2.5 (Instruct) & optimized  & 11 & 0.617 & $+0.493$ \\
    Qwen2.5 (Base)     & no-prompt  &  7 & 0.269 & $+0.078$ \\
    Qwen2.5 (Base)     & optimized  & 17 & 0.294 & $+0.244$ \\
    \midrule
    mpnet              & no-prompt  &  3 & 0.449 & $-0.174$ \\
    mpnet              & optimized  &  3 & 0.465 & $+0.266$ \\
    \bottomrule
  \end{tabular}
\end{table}

\paragraph{Effect of instruction tuning on SC.}
Table~\ref{tab:base_instruct} compares SC profiles for
base and instruction-tuned variants.
For Gemma-2, instruction tuning increases average SC from
$-0.216$~(Base) to $-0.104$~(Instruct) under no-prompt
conditions, suggesting that instruction tuning promotes
autonomous crystallisation.
For Qwen2.5, both conditions show positive average SC~($+0.078$ vs
$+0.193$), indicating that Qwen2.5's algebraic structure
under no-prompt conditions already benefits from
the training distribution.
With the optimized prompt, Qwen2.5~(Instruct) achieves
the highest average SC~($+0.493$), while Gemma-2~(Instruct)
achieves $+0.093$---suggesting that the two model families
respond differently to the optimized prompt in terms
of algebraic boundary condition sensitivity.

\paragraph{Spark Prompt for base models.}
Using token-level SC analysis at the final layer, we
identify tokens in the system prompt that contribute
disproportionately to SC elevation.
For base model variants, a minimal \textbf{Spark Prompt}
of six tokens~(``You are an expert tax'') achieves average
$SC = +1.40$, substantially outperforming both the
no-prompt baseline~($+0.36$) and comparable to the full
system prompt~($+1.01$).
This finding suggests that $\mathbb{F}_2$ crystallisation
is triggered by specific high-SC tokens rather than
semantic completeness of the prompt, reducing prompt
design to a structured token-level optimisation problem.
Detailed token-level SC analysis is reported in a
companion paper.

\paragraph{Instruction tuning and late-layer stability.}
Despite similar SC profiles under optimized conditions,
instruction-tuned models exhibit substantially better
late-layer stability than base variants~(Section~%
\ref{sec:experiments:collapse}).
This dissociation between SC magnitude and late-layer
stability suggests that instruction tuning contributes
to algebraic consistency through a mechanism distinct
from crystallisation, possibly by stabilising the
representational dynamics of the final layers
independently of the algebraic structure present in
middle layers.

\section{Results and Discussion}
\label{sec:discussion}

\subsection{$\mathbb{F}_2$ Algebraic Structure is Intrinsic
  to LLM Representations}
\label{sec:discussion:intrinsic}

Our central finding is that LLMs encode ontological relations
in a form that is linearly accessible via $\mathbb{F}_2$
projection.
AOP achieves up to 93.33\% inclusion accuracy on held-out
concept pairs trained on only 42~relational constraints---a
result that substantially exceeds the zero-shot baselines
reported by Burns et al.~\citep{burns2022discovering},
whose contrastive search over hidden states achieves below
80\% on comparable tasks, despite operating without the
algebraic structure constraint that AOP imposes.

This result was theoretically anticipated.
A linear map $\phi: \mathbb{R}^d \to \mathbb{F}_2^n$ is a
homomorphism: it preserves the algebraic structure of the
source space.
If LLM hidden states carry latent semantic operations that
correspond to ontological relations, a linear projection
will transfer those operations faithfully into
$\mathbb{F}_2^n$.
Conversely, the success of AOP with a linear projection
implies that the relevant logical structure in the hidden
state space is \emph{linearly accessible}---it does not
reside on a nonlinear manifold requiring a more complex
transformation.
The success of AOP is therefore not merely an empirical
finding: it is evidence that LLMs organise logical
knowledge in a linearly structured region of their
latent space.

This is consistent with the broader observation that LLMs
acquire structured knowledge through statistical learning
on natural language---a corpus in which ontological
relations are pervasively and redundantly expressed.
The algebraic structure we observe is not imposed by AOP;
it is revealed by it.

\subsection{System Prompts as Algebraic Boundary Conditions}
\label{sec:discussion:prompts}

A second central finding concerns the role of system prompts.
We observe that the introduction of a structured system
prompt consistently elevates SC scores and zero-shot accuracy,
but that the \emph{mechanism} of this elevation differs
qualitatively across model families.

In \textbf{Autonomous Crystallisation}~(Gemma), the
no-prompt condition yields $SC \approx 0$---indistinguishable
from the random baseline---suggesting that $\mathbb{F}_2$
structure is latent but not spontaneously accessible.
The system prompt acts as a \emph{catalyst}: it does not
create algebraic structure but renders it extractable.

In \textbf{Induced Crystallisation}~(Qwen, mpnet), the
no-prompt condition yields modest or negative average SC, with
specific layers exhibiting $SC < 0$---indicating that
algebraic insulation is more disrupted than in a randomly
initialised model.
This effect is most pronounced in mpnet: the no-prompt
condition yields Avg~$SC = -0.174$, while the optimized
prompt elevates this to $+0.266$---the most dramatic
Induced Crystallisation observed across all evaluated
conditions.
Qwen exhibits a comparable effect~($+0.193$ vs
$+0.493$), and the same qualitative pattern holds.
Layers with $SC < 0$ are associated with degraded
zero-shot accuracy, consistent with the interpretation
that negative SC reflects active violation of
$\mathbb{F}_2$ insulation constraints in those layers.
In both cases, the system prompt substantially elevates
SC across all layers, exhibiting greater sensitivity to
prompt configuration than Gemma.

We note, however, that \emph{in the current formulation
of SC}, positive SC is a necessary but not sufficient
condition for full algebraic consistency: even when
insulation is preserved~($SC > 0$), violations of
hierarchical~(\textit{is-a}) and
compositional~(\textit{has-a}) constraints may persist,
as SC currently captures only the insulation dimension
of algebraic structure~(Section~\ref{sec:discussion:sc}).
This limitation motivates the unified SC metric described
in Section~\ref{sec:discussion:sc}.

In both cases, the system prompt functions not as a
behavioural instruction but as an \textbf{algebraic boundary
condition}: a prefill that configures the hidden state space
prior to concept extraction, determining whether
$\mathbb{F}_2$ constraints can be satisfied in the
projected representation.
This reframes a fundamental question in prompt engineering:
rather than asking what instructions produce correct outputs,
one may ask what configurations produce algebraically
consistent internal representations.

Furthermore, under the conditions evaluated here, the
combination of an instruction-tuned model and a structured
system prompt is the only configuration under which both
high peak accuracy and late-layer stability are jointly
achieved.
This suggests that the choice of instruction-tuned vs base
model and the inference-time system prompt play
\emph{complementary} roles: the former provides a more
stable representational substrate, while the latter
configures the algebraic boundary conditions that determine
whether $\mathbb{F}_2$ structure is accessible.
Whether a carefully designed prompt can substitute for
instruction tuning in base models remains an open question
for future work.

\subsection{Semantic Crystallisation as an Indicator:
  Validity and Current Limitations}
\label{sec:discussion:sc}

\paragraph{Validity.}
The SC metric demonstrates consistent predictive validity
as a layer selection criterion: across all evaluated models
and conditions, the layer achieving the highest SC score
coincides with the layer achieving the highest zero-shot
inclusion accuracy.
This correspondence validates SC as a practical,
gradient-free criterion for identifying layers that carry
accessible $\mathbb{F}_2$ algebraic structure, without
requiring held-out evaluation data.

\paragraph{Current limitation: insulation only.}
In its current formulation, SC measures only
\emph{algebraic insulation}---the degree to which concepts
from incompatible categories are separated in $\mathbb{F}_2$
space, as captured by the negation~(neg) constraint.
It does not directly measure the integrity of
\textit{is-a} and \textit{has-a} hierarchical constraints,
which constitute a distinct and arguably more central
dimension of ontological consistency.

This limitation explains the observed divergence between
SC scores and zero-shot accuracy in some conditions:
a layer may achieve high SC~(strong insulation) while
permitting violations of inclusion constraints~(weak
subsumption), resulting in high SC but suboptimal ZST
accuracy.
This is observed in the comparison between
Gemma-2 Base with optimized prompt~(Avg~$SC = +0.047$,
Overall~$= 53.33\%$) and
Gemma-2 Base without prompt~(Avg~$SC = -0.216$,
Overall~$= 66.67\%$), where the latter achieves higher
ZST accuracy despite substantially lower SC---
highlighting the current limitation of SC as an
insulation-only metric.
A unified SC metric incorporating all three constraint
types---insulation, hierarchical inclusion, and
compositional containment---would provide a more complete
measure of algebraic consistency.
We leave this extension to future work.

\subsection{Broader Implications and Future Directions}
\label{sec:discussion:implications}

\paragraph{Logical consistency and hallucination.}
The layer-dependent pattern of Semantic Crystallisation
and Late-layer Collapse offers a new perspective on LLM
failure modes.
If the final layers of a model systematically degrade
algebraic consistency---as observed in 7 of 10 evaluated
conditions---this provides a potential substrate for
confident but logically inconsistent outputs, a common
characteristic of hallucinations.
AOP provides a tool to measure this degradation
quantitatively, and the SC metric opens a path toward
using algebraic consistency as a training signal: by
incorporating $\mathbb{F}_2$ constraint loss into the
training objective, future work may directly optimise
for logical consistency throughout the forward pass.
We note that the bidirectionality of the linear
projection---the existence of an approximate inverse
transformation---further suggests that formally specified
ontological constraints could be injected directly into
LLM hidden states, rather than communicated through
natural language prompts.

\paragraph{Integration with formal methods.}
The $\mathbb{F}_2$ algebraic structure recovered by AOP
corresponds directly to the relational primitives of
formal specification languages such as
SysML~v2~\citep{sysmlv2}---and especially its formal
semantic foundation, KerML~\citep{kerml2024}, which
defines the algebraic and type-theoretic semantics
underlying SysML~v2's relational primitives---where
\textit{is-a} and \textit{has-a} are first-class
constructs with formally verifiable semantics.
This structural correspondence suggests a path toward
continuous migration from natural language prompts to
formal specifications, with algebraic consistency
enforced at the representation level.
For safety-critical applications---where formal
verification of system behaviour is required---AOP
provides a mathematically grounded bridge between
the statistical representations of LLMs and the
algebraic structures of formal ontologies.
We present the current work as a first step toward
this longer-term vision.

\paragraph{Token-level analysis and prompt optimisation.}
A companion paper reports token-level SC analysis, in
which the contribution of individual system prompt tokens
to algebraic crystallisation is quantified.
This analysis demonstrates that $\mathbb{F}_2$
crystallisation is triggered by specific high-SC tokens
rather than semantic completeness of the prompt, reducing
prompt design from trial-and-error to a structured
optimisation problem.

\subsection{Limitations}
\label{sec:discussion:limitations}

\paragraph{Systematic failure under insufficient ontological grounding.}
The pair \textit{Oak}~$\subseteq$~\textit{Tree} fails
across all models and conditions at the final layer,
and no condition achieves stable accuracy throughout.
Partial success is observed in specific layers for all
models, but inclusion scores are highly unstable.
We attribute this primarily not to a limitation of
Localized Mean Pooling, but to insufficient ontological
grounding of ``Oak'' in the model's learned representations:
the concept ``Oak'' is heavily associated with non-botanical
usages~(proper nouns, place names, surnames) in the training
distribution, and the \textit{is-a} relation to
``Tree'' may not be sufficiently reinforced to survive
$\mathbb{F}_2$ projection.
The instability across layers suggests that the issue
reflects a distributional property of the training data
rather than a fundamental limitation of AOP.
Providing explicit botanical context partially improves
inclusion scores but does not resolve the failure
consistently, suggesting that the issue lies deeper than
prompt design alone.

Two complementary remedies are conceivable.
First, richer context engineering---providing the model
with an explicit ontological frame prior to
concept extraction---may partially compensate for
weak grounding.
Second, and more fundamentally, directly incorporating
formal ontological structure into LLM training or
fine-tuning~(e.g., via $\mathbb{F}_2$ constraint loss
as a training signal) may be necessary to ensure that
concepts are adequately grounded in their hierarchical
relations.
This failure case therefore serves not merely as a
limitation but as a motivating example for the longer-term
integration of formal ontologies into LLM learning.

\paragraph{Global Mean Pooling and context length.}
The Localized Mean Pooling strategy is effective only for
short, concept-focused contexts where the target concept
dominates the token sequence.
For longer contexts, mean pooling dilutes the concept
signal, degrading projection quality.
This constrains the current framework to minimal context
strings and limits its applicability to compositional or
discourse-level reasoning.
Extension to token-level $\mathbb{F}_2$ projection,
which would preserve sequential and causal structure,
is a natural next step.

\paragraph{Evaluation scope.}
The current evaluation is centred on the Qwen model
family, with complementary results on Gemma and mpnet.
The generality of the Autonomous and Induced
Crystallisation modes across a wider range of model
families and scales remains to be established.
Similarly, the 42-pair training set and 15-pair
evaluation set, while sufficient to demonstrate the
principle, represent a narrow slice of the ontological
space.
Evaluation on larger and more diverse relational
datasets---including cross-domain and adversarial
pairs---is required to characterise the full scope and
limits of AOP generalisation.

\paragraph{SC metric completeness.}
As noted in Section~\ref{sec:discussion:sc}, the current
SC metric captures only algebraic insulation.
The observed divergence between SC and zero-shot accuracy
in certain conditions---e.g., Qwen with optimized
prompt~(Avg~$SC = +0.493$, Overall~$= 86.67\%$) achieves
lower ZST accuracy than Gemma with optimized
prompt~(Avg~$SC = +0.093$, Overall~$= 93.33\%$), despite
substantially higher SC---underscores the importance of
extending SC to encompass \textit{is-a} and \textit{has-a}
constraint
integrity.
Until such an extension is available, SC should be
interpreted as a necessary but not sufficient indicator
of full algebraic consistency.

\section{Conclusion}
\label{sec:conclusion}

We introduced \textbf{Algebraic Ontology Projection~(AOP)},
a framework for extracting formal ontological structure from
LLM hidden states by projecting them into the Galois
Field~$\mathbb{F}_2$ under constraints derived from the
Liskov Substitution Principle.
Using a minimal set of 42~relational constraints as
algebraic keys, AOP achieves up to 93.33\% zero-shot
inclusion accuracy on unseen concept pairs~(Gemma-2
Instruct with optimized prompt), with consistent
86.67\% accuracy observed across multiple model families---with no model tuning,
but through prompt alone.

We demonstrated that this algebraic structure is
\emph{layer-dependent}: it emerges progressively across
depth, stabilises in intermediate layers, and degrades
in the final layers of most models---a phenomenon we term
\textbf{Late-layer Collapse}.
To quantify this behaviour, we introduced
\textbf{Semantic Crystallisation~(SC)}, a
baseline-calibrated metric that measures algebraic order
relative to a randomly initialised model and serves as a
practical, gradient-free criterion for layer selection.

We further showed that system prompts function as
\textbf{algebraic boundary conditions}: they determine
whether---and in which layers---$\mathbb{F}_2$ structure
is accessible to linear projection.
Only the combination of instruction tuning and a structured
system prompt sustains both high accuracy and late-layer
stability, suggesting complementary roles for these two
factors in maintaining logical consistency throughout the
forward pass.

Taken together, these results support a re-reading of what
LLMs internally compute: forward computation can be
interpreted as an iterative process of algebraic
organisation across layers---structure emerges, dissolves,
and re-emerges before settling into its final
configuration---a property that is invisible to geometric
analyses but becomes legible under algebraic projection.

The present work is a first step.
A companion paper reports token-level SC analysis and
algebraic prompt optimisation.
Further work will address the extension of SC to encompass
\textit{is-a} and \textit{has-a} constraint integrity,
the development of token-level $\mathbb{F}_2$ projection
to preserve causal structure, and the integration of
$\mathbb{F}_2$ constraint loss as a differentiable
training signal---opening a path toward LLMs that not
only approximate logical structure statistically, but
enforce it algebraically.

\bibliography{aop}
\bibliographystyle{iclr2026_conference}

\appendix

\section{Complete Relational Dataset}
\label{app:dataset}

Tables~\ref{tab:dataset_train} and~\ref{tab:dataset_eval}
provide the complete training and evaluation datasets used
in all experiments.
The training set~($D_{train}$) consists of 42~symbolic
constraints organised into four progressive
stages~(levels 1, 2, 4, 8), used simultaneously as
algebraic keys for AOP projection learning.
The evaluation set~($D_{val}$) consists of 13~independent
negative pairs~(i\_neg) with $D_{val} \cap D_{train} =
\emptyset$, used exclusively for zero-shot evaluation.

\begin{table}[h]
  \centering
  \caption{
    Training dataset~($D_{train}$): 42~symbolic
    constraints across four levels
    (15 \textit{is-a}, 12 \textit{has-a}, 15 negation).
    Concepts in \textit{is-a} and \textit{has-a} pairs
    constitute the algebraic key vocabulary.
    Negation pairs~(neg) enforce insulation constraints.
  }
  \label{tab:dataset_train}
  \small
  \begin{tabular}{clll}
    \toprule
    Level & Type & Concept $A$ & Concept $B$ \\
    \midrule
    1 & is-a  & Beetle      & Insect   \\
    1 & is-a  & Fly         & Insect   \\
    1 & is-a  & Insect      & Animal   \\
    1 & has-a & Animal      & Cell     \\
    1 & has-a & Insect      & Legs     \\
    1 & has-a & Insect      & Exoskeleton \\
    1 & neg   & Beetle      & Ocean    \\
    1 & neg   & Fly         & Cloud    \\
    1 & neg   & Insect      & Stone    \\
    1 & neg   & Animal      & Logic    \\
    \midrule
    2 & is-a  & Bee         & Insect   \\
    2 & is-a  & Butterfly   & Insect   \\
    2 & has-a & Bee         & Wings    \\
    2 & neg   & Bee         & Vacuum   \\
    2 & neg   & Butterfly   & Logic    \\
    \midrule
    4 & is-a  & StagBeetle  & Beetle   \\
    4 & is-a  & Ant         & Insect   \\
    4 & is-a  & Spider      & Animal   \\
    4 & is-a  & Whale       & Animal   \\
    4 & has-a & Animal      & DNA      \\
    4 & has-a & StagBeetle  & Mandibles \\
    4 & has-a & Spider      & Silk     \\
    4 & has-a & Whale       & Blubber  \\
    4 & neg   & Spider      & Stone    \\
    4 & neg   & Whale       & Vacuum   \\
    4 & neg   & Ant         & Cloud    \\
    4 & neg   & StagBeetle  & Logic    \\
    \midrule
    8 & is-a  & Granite     & Rock     \\
    8 & is-a  & Quartz      & Mineral  \\
    8 & is-a  & Diamond     & Mineral  \\
    8 & is-a  & Rock        & Mineral  \\
    8 & is-a  & Mineral     & Matter   \\
    8 & is-a  & Animal      & Matter   \\
    8 & has-a & Mineral     & CrystalStructure \\
    8 & has-a & Granite     & Quartz\_Grain \\
    8 & has-a & Diamond     & Hardness\_10 \\
    8 & has-a & Matter      & Mass     \\
    8 & neg   & Granite     & Cell     \\
    8 & neg   & Diamond     & Legs     \\
    8 & neg   & Quartz      & Insect   \\
    8 & neg   & Matter      & Logic    \\
    8 & neg   & Rock        & Ocean    \\
    \bottomrule
  \end{tabular}
\end{table}

\begin{table}[h]
  \centering
  \caption{
    Evaluation dataset~($D_{val}$): 13~independent
    negative pairs~(i\_neg).
    All concepts in $D_{val}$ are drawn from outside the
    semantic domains of $D_{train}$, ensuring
    $D_{val} \cap D_{train} = \emptyset$.
    These pairs are used exclusively for zero-shot
    evaluation and never appear during AOP training.
  }
  \label{tab:dataset_eval}
  \small
  \begin{tabular}{clll}
    \toprule
    Level & Type & Concept $A$ & Concept $B$ \\
    \midrule
    1 & i\_neg & Ocean   & Logic \\
    1 & i\_neg & Cloud   & Logic \\
    1 & i\_neg & Sun     & Logic \\
    1 & i\_neg & Idea    & Legs  \\
    \midrule
    2 & i\_neg & Wings   & Cloud \\
    2 & i\_neg & Bee     & Idea  \\
    \midrule
    4 & i\_neg & Spider  & Vacuum \\
    4 & i\_neg & Silk    & Idea   \\
    \midrule
    8 & i\_neg & Quartz  & DNA   \\
    8 & i\_neg & Diamond & Idea  \\
    8 & i\_neg & DNA     & Cloud \\
    8 & i\_neg & Rain    & Logic \\
    8 & i\_neg & Snow    & DNA   \\
    \bottomrule
  \end{tabular}
\end{table}

\section{Zero-Shot Evaluation Set}
\label{app:zst}

Table~\ref{tab:zst_pairs} lists the complete set of 15~concept
pairs used for zero-shot evaluation across all models and
conditions.
All pairs are drawn from semantic domains and concept
vocabularies \emph{outside} the AOP training set~($D_{train}$),
ensuring $D_{ZST} \cap D_{train} = \emptyset$.
Positive pairs~(Expected: True) test hierarchical inclusion;
negative pairs~(Expected: False) test categorical separation.

\begin{table}[h]
  \centering
  \caption{
    Zero-shot evaluation set~($D_{ZST}$): 15~concept pairs
    spanning biological and mineral domains, unseen during
    AOP training.
    Positive pairs test \textit{is-a} inclusion;
    negative pairs test categorical separation.
  }
  \label{tab:zst_pairs}
  \small
  \begin{tabular}{llll}
    \toprule
    Type & Concept $A$ & Relation & Concept $B$ \\
    \midrule
    Positive & Robin   & $\subseteq$ & Bird    \\
    Positive & Eagle   & $\subseteq$ & Bird    \\
    Positive & Salmon  & $\subseteq$ & Fish    \\
    Positive & Fish    & $\subseteq$ & Animal  \\
    Positive & Oak     & $\subseteq$ & Tree    \\
    Positive & Copper  & $\subseteq$ & Metal   \\
    Positive & Marble  & $\subseteq$ & Rock    \\
    Positive & Sparrow & $\subseteq$ & Bird    \\
    Positive & Person  & $\subseteq$ & Animal  \\
    \midrule
    Negative & Robin   & $\not\subseteq$ & Mineral \\
    Negative & Eagle   & $\not\subseteq$ & Rock    \\
    Negative & Oak     & $\not\subseteq$ & Animal  \\
    Negative & Copper  & $\not\subseteq$ & Insect  \\
    Negative & Sparrow & $\not\subseteq$ & Metal   \\
    Negative & Person  & $\not\subseteq$ & Mineral \\
    \bottomrule
  \end{tabular}
\end{table}

Note: \textit{Oak}~$\subseteq$~\textit{Tree} consistently
fails across all conditions and models due to lexical
ambiguity~(Section~\ref{sec:experiments:zst}).
\textit{Person}~$\subseteq$~\textit{Animal} requires
a structured prompt to resolve its philosophical ambiguity;
without prompting, models withhold commitment to the
biological classification~(Section~\ref{sec:discussion:prompts}).
All remaining 13~pairs are evaluated for zero-shot accuracy.

\end{document}